\documentclass{article}
\usepackage[utf8]{inputenc}
\usepackage{amsfonts,amsmath}
\usepackage{mdframed}
\usepackage{graphicx}
\usepackage{hyperref}

\usepackage{multirow}
\usepackage{array}
\usepackage{subcaption}

\usepackage{url}

\usepackage{tikz}
\usetikzlibrary{arrows.meta,,positioning}

\usepackage{caption}

\usepackage[ruled,boxed,linesnumbered,commentsnumbered,longend]{algorithm2e}

\usepackage{amsthm}
\theoremstyle{definition}
\newtheorem{prop}{Proposition}

\usepackage{multirow}
\usepackage{breqn}

\def\old#1{}

\def\ol{\overline}

\def\tl{\tilde}

\def\p{\pi}

\def\m{\mu}

\def\frac#1#2{{#1\over #2}}

\def\ol#1{\overline{#1}}

\old{

}

\title{Most Likely Sequence Generation for $n$-Grams, Transformers, HMMs, and Markov Chains, by Using Rollout Algorithms\footnote{
This work was carried out at the Fulton School of Computing, and Augmented Intelligence, Arizona State University, Tempe, AZ.}}

\author{Yuchao Li and Dimitri Bertsekas\\
School of Computing, and Augmented Intelligence,\\ Arizona State University, Tempe, AZ.}
\date{}
\usepackage{setspace}
\begin{document}
\maketitle

\begin{abstract}
In this paper we consider a transformer with an $n$-gram structure, such as the one underlying ChatGPT. The transformer provides next word probabilities, which can be used to generate word sequences. We consider methods for computing word sequences that are highly likely, based on these probabilities. Computing the optimal (i.e., most likely) word sequence starting with a given initial state is an intractable problem, so we propose methods to compute highly likely sequences of $N$ words in time that is a low order polynomial in $N$ and in the vocabulary size of the $n$-gram. These methods are based on the rollout approach from approximate dynamic programming, a form of single policy iteration, which can  improve the performance of any given heuristic policy. In our case we use a greedy heuristic that generates as next word  one that has the highest probability. We show with analysis, examples, and computational experimentation that our methods are capable of generating highly likely sequences with a modest increase in computation over the greedy heuristic.  While our analysis and experiments are focused on Markov chains of the type arising in transformer and ChatGPT-like models, our methods apply to general finite-state Markov chains, and related inference applications of Hidden Markov Models (HMM), where Viterbi decoding is used extensively.
\end{abstract}


\vspace{1pc}
\section{Introduction}
Generative pre-trained transformers (GPT) have sparked a lot of enthusiasm for innovative applications in many problem domains, aided by powerful openly available software, and easy-to-use natural language interfaces. At the same time, transformers have been established as a flexible and powerful model, which generalizes in important ways earlier forms of neural networks by using the attention mechanism and more complex nonlinearities (see the recent textbook by Bishop and Bishop [BiB24], Chapter 12, for a  description of the transformer architecture, with earlier references to the literature).  

\begin{figure}[ht]
\captionsetup{singlelinecheck=off}
\begin{center}
\centerline{{\includegraphics[width=1\columnwidth]{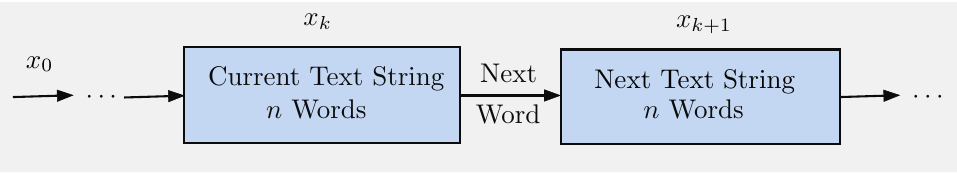}}}
\caption{\small Schematic visualization of an $n$-gram. Given the ($n$-word)  text string $x_k$ generated at time $k$, it generates the next  ($n$-word) text string $x_{k+1}$ by  adding a word at the front end of $x_k$, and deleting the word at the back end of $x_k$.}
		\label{figdynamic}
		\end{center}
\end{figure}

In this paper, we will view a transformer in terms of the classical $n$-gram model that generates a sequence $\{x_1,\ldots,x_N\}$ of text strings, starting from some initial string $x_0$. Each string $x_k$ consists of a sequence of $n$ words, chosen from a given list (the {\it vocabulary} of the $n$-gram). The $k$th string $x_k$  is transformed into the next  string  $x_{k+1}$ by adding a word at the front end of $x_k$ and deleting the word at the back end of $x_k$; see Fig.\ \ref{figdynamic}. Here $n$ and $N$ are fixed positive integers. We refer to the book by Jurafsky and Martin [JuM23], Chapter 3, for description and applications of $n$-grams and their connection to transformers, including historical and other references. 

Given a text string $x_k$, the $n$-gram provides probabilities $p(x_{k+1}\mid x_k)$ for the next text string $x_{k+1}$. These probabilities also define the  probabilities of the possible next words, since $x_{k+1}$ is determined by the next word that is added to the front of $x_k$. We assume that the probabilities $p(x_{k+1}\mid x_k)$ depend only on $x_k$. Thus they can be viewed as the transition probabilities of a stationary Markov chain, whose state space is the set of all  $n$-word sequences $x_k$\footnote{The stationarity assumption simplifies our notation, but is not essential to our methodology, as we will discuss later.} Bearing this context in mind, we will also refer to $x_k$ as the {\it state} (of the underlying Markov chain).

The probabilities $p(x_{k+1}\mid x_k)$ can provide guidance for generating state sequences with some specific purpose in mind. To this end, a transformer may use a (next word) {\it selection policy\/},  i.e., a (possibly time-dependent) function $\m_k$, which  selects the text string that follows $x_k$ as
$$x_{k+1}=\m_k(x_k).$$
We are generally interested in selection policies that give preference to high probability future words. Two frequently considered policies are:

\begin{itemize}
\item[$\bullet$] {\it Greedy selection\/}: Here, given $x_k$, the next state $x_{k+1}$ is selected to be the one that maximizes $p(x_{k+1}\mid x_k)$:
$$\m_k(x_k)\in\arg\max_{x_{k+1}}p(x_{k+1}\mid x_k).$$
To generate a sequence $\{x_1,\ldots,x_N\}$ of $N$ states, this selection policy requires computation that is proportional to $N$ and to the size of the $n$-gram's vocabulary (the number of all possible next words).

\item[$\bullet$] {\it Most likely sequence selection\/}: Here, we consider all the possible sequences $\{x_1,\ldots,x_{N}\}$ that can be generated starting with $x_0$, and select the sequence that is most likely, i.e.,  has maximal probability of occurrence. This selection policy requires computation that grows linearly with $N$ and exponentially with the size  of the $n$-gram's vocabulary.
\end{itemize}

The next word selection policy affects substantially the behavior of the $n$-gram, depending on the practical context at hand. In particular, contrary to the greedy selection method, the most likely sequence selection method {\it takes into account future selections\/}, beyond the next word choice. Unfortunately, however, computing the most likely sequence is intractable as noted earlier.  It can only be obtained by generating the tree of the possible sequences $\{x_1,\ldots,x_{N}\}$ given the initial state $x_0$, and then using a shortest path-type  method to compute the most likely sequence. For example, forward or backward dynamic programming (DP) can be used; see the next section.

In this paper, we propose an intermediate next word selection method, the {\it rollout selection policy\/}, which is an approximate DP method, well known for its simplicity and good performance record, owing to its close connection to the fundamental DP algorithm of policy iteration. The rollout approach was first applied to deterministic combinatorial optimization problems in the paper [BTW97], and it has been extensively investigated and tested in the  context of many types of DP problems, both deterministic and stochastic (see the second author's textbooks [Ber19], [Ber20], [Ber23], which provide extensive references to earlier research). 

The rollout approach produces highly likely (near optimal) sequences, with computation that is larger than the greedy selection method by a factor that is proportional to $N$ and to the size of the $n$-gram's vocabulary. This represents a substantial increase over the greedy selection method, but is still far lower than the exponential computation of the most likely selection method. Let us also note that variants that aim to reduce further the computational requirements of the rollout selection policy are possible, including simplified and truncated versions, which will be discussed later. 

The good performance of the rollout algorithm and its variants owes its success and reliability to its connection with the approximation in value space approach of reinforcement learning. In this context, the rollout algorithm and its approximate or enhanced variants are interpreted as a step of Newton's method for solving the Bellman equation underlying the corresponding DP problem; this is one of the major conceptual focal points of the books [Ber20], [Ber22], [Ber23].

In the next section we will present the greedy, most likely, and rollout selection policies within a more general context where the transformer is replaced by an arbitrary stationary finite-state Markov chain. In Section 3, we will discuss variants of the rollout approach, including simplified, truncated, multistep, and multi-iteration rollout. In Section 4, we will compare analytically the three types of policies. Finally, in Section 5, we will provide results of our computational experimentation.

\section{The Greedy, Most Likely, and Rollout Selection Methods for a Stationary Markov Chain}

In this section we will formally describe the greedy, most likely, and rollout selection policies within a general Markov chain framework. In particular, we consider a stationary Markov chain with a finite state space $X$. We will generally use the symbols $x$ and $y$ for states, and we will denote the chain's transition probabilities by $p(y\mid x)$. We assume that given a state $x$, the probabilities $p(y\mid x)$ are either known or can be generated on-line by means of software such as a transformer. 

We assume stationarity of the Markov chain in part to alleviate an overburdened notation, and also because $n$-gram and transformer models are typically assumed to be stationary. However, {\it the rollout methodology and the manner in which we use it do not depend at all on stationarity} of the transition probabilities, or infinite horizon properties of Markov chains, such as ergodic classes, transient states, etc. In fact, they also do not depend on the stationarity of the state space either. Only the Markov property is used in our discussion, i.e., the probability of the next state depends on the immediately preceding state, and not on earlier states.

A {\it selection policy} $\p$ is a sequence of functions $\{\m_0,\ldots,\m_{N-1}\}$, which given the current state $x_k$, determines the next state $x_{k+1}$ as 
$$x_{k+1}=\m_k(x_k).$$
Note that for a given $\p$, the state evolution is {\it deterministic\/}; so for a given $\p$ and $x_0$, the generated state sequence $\{x_1,\ldots,x_N\}$ is fully determined. Moreover the choice of $\p$ is arbitrary, although we are primarily interested in policies $\p$ that give preference to high probability next states.

\begin{figure}[ht]
\captionsetup{singlelinecheck=off}
\begin{center}
\centerline{{\includegraphics[width=1\columnwidth]{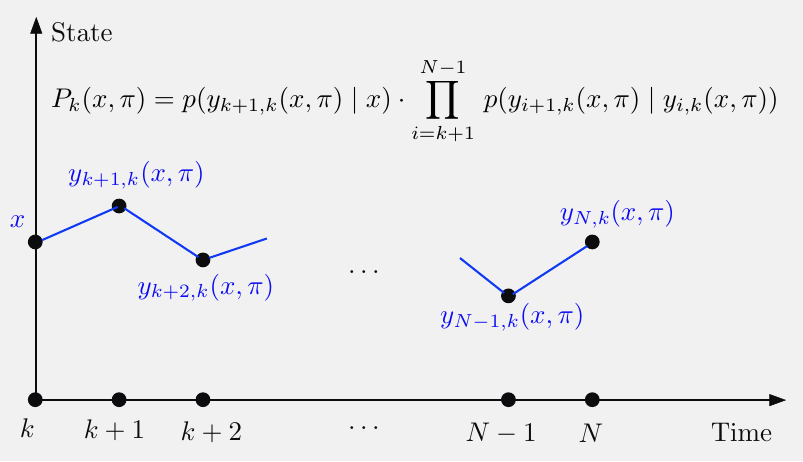}}}
\caption{\small Illustration of the state trajectory generated by a policy $\p$, starting at state $x$ at time $k$.  The probability of its occurrence, $P_k(x,\p)$, is the product of the transition probabilities along the $N-k$ steps of the trajectory [cf.\ Eq.\ (\ref{multrule})].
}
		\label{figtrajectory}
		\end{center}
\end{figure}

Given a policy $\p=\{\m_0,\ldots,\m_{N-1}\}$ and a starting state $x$ at time $k$, the state at future times $m>k$ is denoted by $y_{m,k}(x,\p)$:
$$y_{m,k}(x,\p)=\hbox{ state at time $m>k$ starting at state $x$ and using $\p$}.$$
The {\it state trajectory generated by a policy $\p$, starting at state $x$ at time $k$\/}, is the sequence 
$$y_{k+1,k}(x,\p),\ldots,y_{N,k}(x,\p),$$
(cf.\ Fig.\ \ref{figtrajectory}), and the probability of its occurrence in the given Markov chain is 
\begin{equation} \label{multrule}
P_k(x,\p)=p\big(y_{k+1,k}(x,\p)\mid x\big)\cdot\prod_{i=k+1}^{N-1}\,p\big(y_{i+1,k}(x,\p)\mid y_{i,k}(x,\p)\big),
\end{equation}
according to the multiplication rule for conditional probabilities.

\subsection*{Optimal/Most Likely Selection Policy}

The {\it most likely selection policy\/}, denoted by $\p^*=\{\m_0^*,\ldots,\m_{N-1}^*\}$, maximizes  over all policies $\p$ the probabilities $P_k(x,\p)$ for every initial state $x$ and time $k$. The corresponding probabilities of $\p^*$, starting at state $x$ at time $k$, are denoted by $P_k^*(x)$:
$$P^*_k(x)=P_k(x,\p^*)=\max_\p P_k(x,\p).$$
One way to compute the policy $\p^*$ and its probabilities $P_k^*(x)$ is to use the following DP-like algorithm, which operates in two stages:

\begin{itemize}
\item[$\bullet$]We first compute the probabilities $P^*_k(x)$ backwards, for all $x$,  according to
\begin{equation}
P^*_k(x)=\max_y\,p(y\mid x)P^*_{k+1}(y),\qquad k=N-1,\ldots,0,
\label{probmostlikely}
\end{equation}
starting with 
$$P^*_N(x)\equiv 1.$$ 
\item[$\bullet$]We then generate sequentially the selections $x_1^*,\ldots,x^*_N$ of $\p^*$ according to
\begin{equation}
x^*_{k+1}=\m_k^*(x^*_k)\in \arg\max_y\,p(y\mid x^*_k)P^*_{k+1}(y),
\label{mostlikely}
\end{equation}
going forwards starting with $x^*_0=x_0$. 	
\end{itemize}
This algorithm is equivalent  to the usual DP algorithm for multistage additive costs, after we take logarithms of the multiplicative expressions defining the probabilities $P_k(x,\p)$.

We note that the problem of finding the most likely sequence generated by a Markov chain arises in many important contexts, and its solution by DP-like methods is well-known. A major example is inference of the sequence of states of a Hidden Markov Model (HMM), given an associated sequence of observed data. This is the problem where Viterbi decoding [Vit67], [For73], and related algorithms are used widely, and it plays an important role in several diverse fields, such as speech recognition [Rab69], [JuM23], computational linguistics and language translation [JuM23], [MaS99], coding and error correction [PrS01], [PrS08], bioinformatics [Edd96], [DEK98], and others. Compared to these fields, the transformer/$n$-gram context tends to involve Markov chains with an intractably larger state space. Moreover, while approximations  are commonly employed in applications of the Viterbi algorithm to these fields, the rollout approach for approximating most likely sequences has not been considered to our knowledge. A DP-oriented discussion of the Viterbi algorithm and its applications to HMM inference is given in Section 2.2.2 of the textbook [Ber17].

\subsection*{Greedy Policy}

At any given state $x_k$, the {\it greedy policy} produces the next state by maximization of the corresponding transition probability over all $y$:
$$\max_y p(y\mid x_k).$$
We assume that ties in the above maximization are broken according to some prespecified deterministic rule. For example if the states are labeled by distinct integers, one possibility is to specify the greedy selection at $x_k$ as the state $y$ with minimal label, among those that attain the maximum above. Note that the greedy policy is not only deterministic, but it is also stationary (its selections depend only on the current state and not on the time $k$). We will consequently use the notation $\ol \p=\{\ol\m,\ldots,\ol\m\}$ for the greedy policy, where 
\begin{equation}
\ol \m(x_k)\in\arg\max_y p(y\mid x_k),
\label{greedy}
\end{equation}
and $\ol \m(x_k)$ is uniquely defined according to our deterministic convention for breaking ties in the maximization above. 
The corresponding probabilities $P_k(x_k,\ol\p)$ are given by the DP-like algorithm
\begin{equation}
P_k(x,\ol\p)=p\big(\ol\m(x)\mid x\big)P_{k+1}\big(\ol\m(x),\ol\p\big),\qquad k=N-1,\ldots,0,
\label{probgreedy}
\end{equation}
starting with 
$$P_N(x,\ol\p)\equiv 1.$$
Equivalently, we can compute $P_k(x,\ol\p)$ by using a forward multiplication of the transition probabilities along the trajectory generated by the greedy policy, starting from $x$; cf.\ Eq.\ (\ref{multrule}).

Greedy search algorithms are used widely in discrete optimization problems, and their principal  limitation is well-known: they choose the locally optimal next state without considering the impact of this choice on future state selections. The rollout approach, to be discussed next, mitigates this limitation with a mechanism for looking into the future, and balancing the desire for a high-probability next state with the potential undesirability of low-probability future states.

\subsection*{Rollout Policy}

\begin{figure}[ht]
\captionsetup{singlelinecheck=off}
\begin{center}
\centerline{{\includegraphics[width=1\columnwidth]{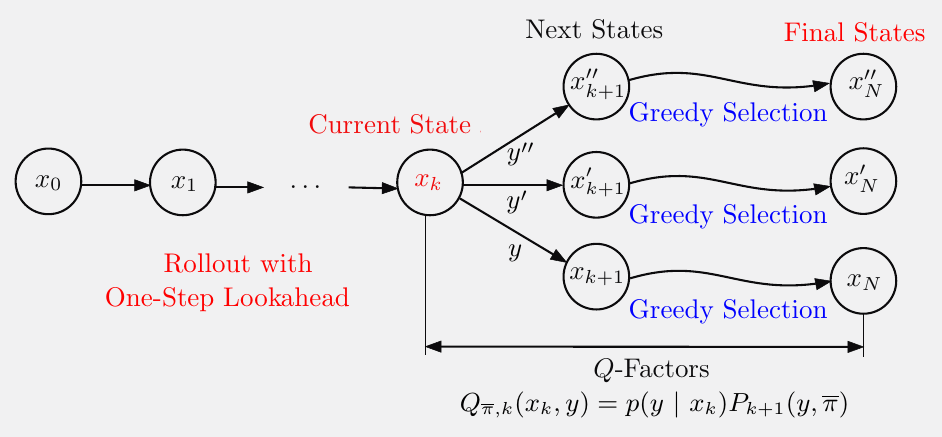}}}
\caption{\small Schematic illustration of the rollout policy with one-step lookahead. At the current state $x_k$, we compute the Q-factors 
$$Q_{\ol\p,k}(x_k,y)=p(y\ |\ x_k)P_{k+1}(y,\ol\p)$$
by running the greedy selection policy from  all possible next states $y$. We then select as next state $x_{k+1}$ the one with maximal Q-factor.
}
		\label{figrollout}
		\end{center}
\end{figure}

At any given state $x_k$, the {\it rollout policy with one-step lookahead} produces the next state, denoted $\tl \m_k(x_k)$,  by maximizing $p(y\mid x_k)P_{k+1}(y,\ol\p)$ over all $y$: 
\begin{equation}
\tl \m_k(x_k)\in\arg\max_y p(y\mid x_k)P_{k+1}(y,\ol\p),
\label{rollout}
\end{equation}
and we will consequently use the notation $\tl \p=\{\tl\m_0,\ldots,\tl\m_{N-1}\}$ for the rollout policy. Thus {\it it optimizes the selection of the first state $y$, assuming that the subsequent states will be chosen using the greedy policy\/}. 
By comparing the maximization (\ref{rollout}) with the one for the most likely selection policy [cf.\ Eq.\ (\ref{mostlikely})], we see that  it chooses the next state similarly, except that $P^*_{k+1}(y)$ (which is hard to compute) is replaced by the (much more easily computable) probability $P_{k+1}(y,\ol\p)$. In particular, the latter probability is computed for every $y$ by running the greedy policy forward starting from $y$ and multiplying the corresponding transition probabilities along the generated state trajectory; see Fig.\ \ref{figrollout}. This is a polynomial computation, which is roughly larger by a factor $q\cdot N$ over the greedy selection method (here $q$ is the number of Q-factors computed at each time step). However, there are ways to reduce this computation, including the use of parallel computation and other possibilities, which we will discuss in Section 3. 

The expression $p(y\mid x_k)P_{k+1}(y,\ol\p)$ that is maximized over $y$ in Eq.\ (\ref{rollout}) is known as the {\it Q-factor of the pair $(x_k,y)$ corresponding to the base policy $\ol \p$\/}, in the terminology of the rollout approach, and is denoted by $Q_{\ol \p,k}(x_k,y)$:  
\begin{equation}
Q_{\ol \p,k}(x_k,y)=p(y\mid x_k)P_{k+1}(y,\ol\p).
\label{qfactor}
\end{equation}
The Q-factor terminology comes from schemes of {\it approximation in value space\/}, which underlie some of the most visible successes of  reinforcement learning; cf.\ the books [Ber22], [Ber23]. In this context, at state $x_k$ we choose the action $y$ that yields the maximal Q-factor.

\subsection*{Rollout Policy with $\ell$-Step Lookahead}

\begin{figure}[ht]
\captionsetup{singlelinecheck=off}
\begin{center}
\centerline{{\includegraphics[width=1\columnwidth]{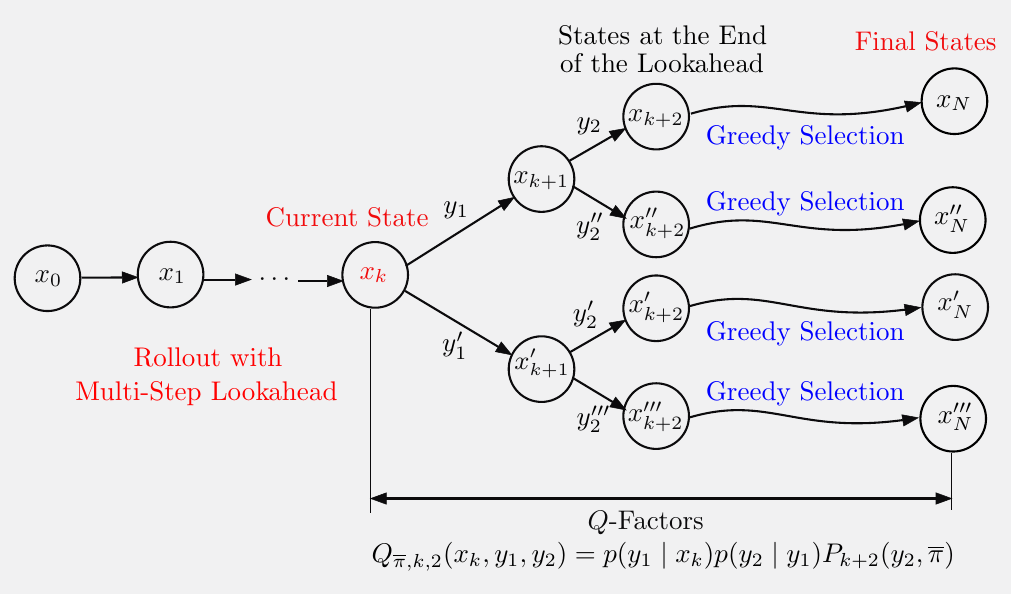}}}
\caption{\small Illustration of $\ell$-step lookahead rollout with $\ell=2$. At the current state $x_k$, we maximize over all pairs $\{y_1,y_2\}$, the $\ell$-step Q-factor 
$$Q_{\ol\p,k,\ell}(x_k,y_1,y_2)=p(y_1\mid x_k)p(y_2\mid y_1)P_{k+\ell}(y_2,\ol\p);$$
cf.\ Eq.\ (\ref{elllook}); the figure illustrates the case $\ell=2$.
If $\{\tl y_1,\tl y_2\}$ is the maximizing sequence, we select $\tl y_1$ and discard $\tl y_2$.}
		\label{figmultistep}
		\end{center}
\end{figure}

Another rollout possibility includes {\it rollout with $\ell$-step lookahead} ($\ell>1$), whereby given $x_k$ we maximize over all sequences $\{y_1,y_2,\ldots,y_\ell\}$ up to $\ell$ steps ahead, the {\it $\ell$-step Q-factor\/}
\begin{equation}
Q_{\ol\p,k,\ell}(x_k,y_1,\ldots,y_\ell)=p(y_1\mid x_k)p(y_2\mid y_1)\cdots p(y_\ell\mid y_{\ell-1}) P_{k+\ell}(y_\ell,\ol\p),
\label{elllook}
\end{equation}
and if $\{\tl y_1,\tl y_2,\ldots,\tl y_\ell\}$ is the maximizing sequence, we select $\tl y_1$ at $x_k$, and discard the remaining states $\tl y_2,\ldots,\tl y_\ell$; see Fig.\ \ref{figmultistep}.\footnote{If $\ell>N-k$, then $\ell$ must be reduced to $N- k$, to take into account end-of-horizon effects.} In practice the performance of $\ell$-step lookahead rollout policies almost always improves with increasing $\ell$. However, artificial examples have been constructed where this not so; see the book [Ber19], Section 2.1.1. Moreover, the computational overhead of $\ell$-step lookahead increases with $\ell$, and for $\ell=N$, the rollout policy coincides with the most likely selection policy.

\subsection*{Illustrative Examples}

\begin{figure}[ht]
\captionsetup{singlelinecheck=off}
\begin{center}
\centerline{{\includegraphics[width=0.5\columnwidth]{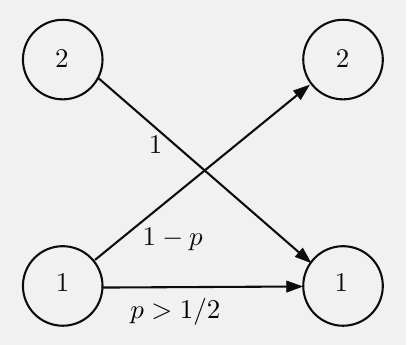}}}
\caption{\small A two-state Markov chain example with transition probabilities as shown next to the transition arcs (the transition not shown in the graph has probability 0). We assume that $x_0=1$, $p>1/2$, and $N$ is even.
}
		\label{figexample1}
		\end{center}
\end{figure}

Let us illustrate the preceding selection policies with examples. 
Figure  \ref{figexample1} provides a two-state example, where the starting state is 1. In this example:

\begin{itemize}
\item[(a)] The greedy policy $\ol \p$ generates the sequence $\{1,1,\ldots,1\}$ and the corresponding probability $P_N(1,\ol\p)$ is equal to $p^N$.
\item[(b)] The most likely selection policy $\p^*$ operates as follows: If $p^2<1-p$, i.e., $0.5<p<0.618$, it generates the sequence $\{1,2,1,2,\ldots,1,2\}$ and the corresponding probability $P_N^*(1)$ is $(1-p)^{N/2}$ (so it is larger than the one of the greedy policy). If $p^2>1-p$, i.e., $0.618<p$, it generates the sequence $\{1,1,\ldots,1\}$ and the corresponding probability $P_N^*(1)$ is $p^N$ (the same as the greedy policy). For $p\approx 1/2$, we have 
$$P_N(1,\ol\p)\approx P_N^*(1)^2,$$
and the greedy policy is far from optimal.
\item[(c)] The rollout policy generates the same sequence as the most likely selection policy. In particular, at state 1 it computes the two Q-factors, corresponding to the next states 1 and 2 [cf.\ Eq.\ (\ref{qfactor})]:
$$Q_{\ol \p,N}(1,1)=p^N,\qquad Q_{\ol \p,N}(1,2)=(1-p)p^{N-2},$$
and selects the action that attains the maximum of the two. This yields the same result as the optimal/most likely selection policy.
\end{itemize}

\begin{figure}[ht]
\captionsetup{singlelinecheck=off}
\begin{center}
\centerline{{\includegraphics[width=0.5\columnwidth]{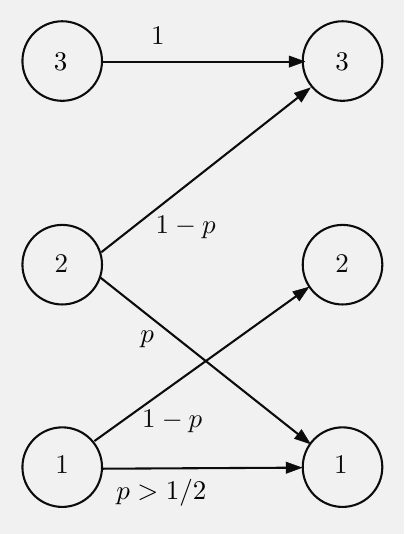}}}
\caption{\small A three-state Markov chain example with transition probabilities as shown next to the transition arcs (the transitions that are not shown in the graph have probability 0). We assume that $x_0=1$ and that $p>1/2$.
}
		\label{figexample2}
		\end{center}
\end{figure}

The preceding example is consistent with general theoretical and empirical results regarding the rollout approach: Its performance is substantially better than the one of its corresponding base policy, and is close to the optimal. Figure  \ref{figexample2} provides a three-state example, which is similar to the two-state example of Fig.\  \ref{figexample1}, and  illustrates the mechanism by which two-step lookahead rollout can work better than the one-step lookahead version. In this example:

\begin{itemize}
\item[(a)] The greedy policy $\ol \p$ generates the sequence $\{1,1,\ldots,1\}$ and the corresponding probability $P_{N}(1,\ol\p)$ is equal to $p^N$.
\item[(b)] The most likely selection policy $\p^*$ generates the sequence $\{1,2,3,3,\ldots,3\}$ and the corresponding probability is $P_{N}^*(1)=(1-p)^{2}$ (much larger than the one of the greedy policy).
\item[(c)] The rollout policy with one-step lookahead at the initial state 1 computes the two Q-factors corresponding to the next states 1 and 2,
$$Q_{\ol \p,N}(1,1)=p^N,\qquad Q_{\ol \p,N}(1,2)=(1-p)p^{N-1}.$$
It thus selects state 1 as next state, and the process is repeated. Thus it generates the sequence $\{1,1,\ldots,1\}$, the same as the greedy policy.
\item[(d)] The rollout policy with two-step lookahead at the initial state 1 computes and compares the two-step ahead Q-factors [cf.\ Eq.\ (\ref{mlook})]. We have:
$$Q_{\ol \p,N}(1,1,1)=p^N,$$
$$Q_{\ol \p,N}(1,1,2)=(1-p)p^{N-1},$$
$$Q_{\ol \p,N}(1,2,1)=(1-p)p^{N-1},$$
$$Q_{\ol \p,N}(1,2,3)=(1-p)^2,$$
so based on the maximizing Q-factor $Q_{\ol \p,N}(1,2,3)$, it selects state 2. Similarly, from state 2 it selects state 3, so the sequence generated is $\{1,2,3,3,\ldots,3\}$, the same sequence as the one obtained by the most likely selection policy. 
\end{itemize}

\section{Variants of the Rollout Policy}

There are several variants of the rollout policy, which are described at length in the literature; cf., the second author's books [Ber19], [Ber20], [Ber22], [Ber23]. We list a few of these variants, which are aimed at either reducing the computational requirements or improving the performance of the rollout approach. In Section 5 we describe these and other additional possibilities in the context of computational experiments.

\subsection*{Simplified Rollout}

A difficulty that arises in the application of rollout is the potentially very large number of the Q-factors that need to be calculated at each time step at the current state $x$ [it is equal to the number of states $y$ for which $p(y\mid x)>0$]. In practice the computation of Q-factors can be restricted to a subset of most probable next states, as per the transition probabilities $p(y\mid x)$ (this is a common expedient in the rollout approach, called {\it simplified rollout\/}; see [Ber20], [Ber22], [Ber23], which also describe conditions under which the performance of the simplified algorithm is not compromised by the simplification). For example, often many of the transition probabilities $p(y\mid x)$ are very close to 0, and can be safely ignored. 

Note that simplified rollout resembles somewhat the method of {\it beam search} for exploring the Markov chain of a large language model (see [JuM23], Section 10.4). However, beam search is quite different, and has the character of pruning the Markov chain starting with the initial state, by discarding the most unlikely next states sequentially over multiple steps. By contrast, simplified rollout reduces the number of calculated Q-factors of the greedy policy at the current step, and does nothing to reduce the calculations in the subsequent steps. 

\subsection*{Truncated Rollout}

Another common way to reduce computation is to truncate the trajectories generated from the next states $y$ by the greedy policy, up to $m$ steps (assuming that $k+m<N$, i.e., if we are more than $m$ steps away from the end of the horizon). In this method, called {\it $m$-step truncated rollout\/}, and discussed extensively in the books [Ber19], [Ber20], [Ber22], and [Ber23], we maximize over $y$ the $m$-step Q-factor of the greedy policy $\ol \p$:
\begin{equation}
p(y\mid x_k)P_{k+1,m}(y,\ol \p),
\label{mlook}
\end{equation}
where
$$P_{k+1,m}(y,\ol \p)=p\big(y_{k+2,k+1}(y,\ol \p)\mid y\big)\cdot\prod_{i=k+2}^{k+m}\, p\big(y_{i+1,k+1}(y,\ol \p)\mid y_{i,k+1}(y,\ol \p)\big)$$
is the  $m$-step product of probabilities along the path generated by the greedy policy $\ol \p$ starting from $y$ at time $k+1$ [cf.\ Eq.\ (\ref{multrule})].
By contrast, in rollout without truncation,  we maximize over $y$  
$$p(y\mid x_k)P_{k+1}(y,\ol\p),$$
where 
$$P_{k+1}(y,\ol\p)=p\big(y_{k+2,k+1}(y,\ol \p)\mid y\big)\cdot\prod_{i=k+2}^{N-1}\, p\big(y_{i+1,k+1}(y,\ol\p)\mid y_{i,k+1}(y,\ol\p)\big)$$
 is the $(N-k-1)$-step product of probabilities along the path generated by the greedy policy starting from $y$ at time $k+1$; cf.\ Eqs.\ (\ref{multrule}) and (\ref{rollout}).

\subsection*{Multiple Policy Iterations - Double Rollout - Complexity Analysis}

Still another possibility is to apply the rollout approach successively, in {\it multiple policy iterations\/}, by using the rollout policy obtained at each iteration as base policy for the next iteration. This corresponds to the fundamental DP algorithm of {\it policy iteration\/}.

Performing on-line just two policy iterations amounts to using the rollout algorithm as a base policy for another rollout algorithm. This has been called {\it double rollout\/}, and it has been discussed in Section 2.3.5 of the  book [Ber20] and Section 6.5 of the book [Ber22]. Generally,  one-step lookahead rollout requires $O(q\cdot N)$ applications of the base policy where $q$ is the number of Q-factors calculated at each time step.\footnote{For a more accurate estimate of the complexity of the greedy, rollout, and double rollout algorithms, note that the basic operation of the greedy operation is the maximization over the $q$ numbers $p(y\mid x_k)$. Thus $m$ steps of the greedy algorithm, as in an $m$-step Q-factor calculation, costs $q\cdot m$ comparisons. In $m$-step truncated rollout, we compare $q$ greedy Q-factors so the number of comparisons per rollout time step is $q^2m+q$. Over $N$ time steps the total is $(q^2m+q)\cdot N$ comparisons, while for the greedy algorithm starting from the initial state $x_0$, the corresponding number is $q\cdot N$. Thus there is an amplification factor of $qm+1$ for the computation of simplified $m$-step truncated rollout over the greedy policy. Similarly it can be estimated that there is an amplification factor of no more than $qm+1$ for using double rollout with (single) rollout as a base policy.}
Thus with each new policy iteration, there is an amplification factor $O(q\cdot N)$ of the computational requirements. Still, however, the multiple iteration approach may be viable, even on-line, when combined with some of the other time-saving computational devices described above (e.g., truncation and simplification to reduce $q$), in view of the relative simplicity of the calculations involved and their suitability for parallel computation. This is particularly so for double rollout. An example where policy iteration has been applied successfully to the game of solitaire is discussed in the paper by Yan et al.\ [YDR04].

The preceding variants of the rollout selection policy will be formalized and compared to the greedy and most likely selection policies, analytically in Section 4, and experimentally in Section 5. In particular, we will show analytically that the  rollout selection policy with one-step lookahead has a {\it performance improvement property\/}: it generates more likely state sequences than the greedy policy, starting from any state. In practice, the improvement is often very substantial, owing to the connection of the method with Newton's method. This has been verified in our computational experiments, and is consistent with the extensive computational experience with rollout algorithms that has been accumulated over 30 years.

\section{Performance Improvement Properties of Rollout Policies}

We will show by induction a performance improvement property of the rollout algorithm with one-step lookahead, namely that for all states $x\in X$ and $k$, we have
\begin{equation}\label{perfimprove}
P_k(x,\ol\p)\le P_k(x,\tl\p),
\end{equation}
i.e., the probability of the sequence generated by the rollout policy is greater or equal to the probability  of the sequence generated by the greedy policy; this is true for any starting state $x$ at any time $k$.

Indeed, for $k=N$ this relation holds, since we have
$$P_N(x,\ol\p)= P_N(x,\tl\p)\equiv1.$$
Assuming that
$$P_{k+1}(x,\ol\p)\le P_{k+1}(x,\tl\p),\qquad\hbox{for all $x$},$$
we will show that
$$P_k(x,\ol\p)\le P_k(x,\tl\p),\qquad\hbox{for all $x$}.$$
Indeed, we use the preceding relations to write
\begin{align}
P_k(x,\tl \p)&=p\big(\tl \m_k(x)\mid x)P_{k+1}\big(\tl \m_k(x),\tl \p\big)\\
& \ge p\big(\tl \m_k(x)\mid x)P_{k+1}\big(\tl \m_k(x),\ol \p\big)\\
& \ge p\big(\ol \m_k(x)\mid x)P_{k+1}\big(\ol \m_k(x),\ol \p\big)\\
&=P_k(x,\ol \p)
\end{align}
where
\begin{itemize}
\item[$\bullet$] The first equality follows from the definition of the probabilities corresponding to the rollout policy $\tl\p$.
\item[$\bullet$] The first inequality follows from the induction hypothesis.
\item[$\bullet$] The second inequality follows from the fact that the rollout choice $\tl \m_k(x)$ maximizes the Q-factor $p\big(y\mid x)P_{k+1}\big(y,\ol \p\big)$ over $y$.
\item[$\bullet$] The second equality follows from the definition of the probabilities corresponding to the greedy policy $\ol\p$.
\end{itemize}
Thus the induction proof of the improvement property (\ref{perfimprove}) is complete.

Clearly, the performance improvement property  continues to hold for double rollout and for successive multiple iterations of the rollout policy, and in fact it can be shown that after  a sufficiently large number of iterations it yields the most likely selection policy. This is a consequence of classical results, which establish the finite convergence to an optimal policy of the policy iteration algorithm for finite-state Markovian decision problems, see e.g., [Ber19].

Performance improvement can also be established for the $\ell$-step lookahead version of the rollout policy, using an induction proof that is similar to the one given above for the one-step lookahead case. Moreover, the books [Ber20], [Ber22], [Ber23] describe conditions under which simplified rollout maintains the performance improvement property. However, it is not necessarily true that the performance of the $\ell$-step lookahead rollout policy improves as $\ell$ increases; see an example in the book [Ber19], Section 2.1.1, and the computational results of the next section. Similarly, it is not necessarily true that the $m$-step truncated rollout policy performs better than the greedy policy.\footnote{It performs better than an $m$-step version of the greedy policy, which generates a sequence of $m+1$ states, starting from the current state and using the greedy policy.}  On the other hand, known performance deterioration examples of this type are artificial and are apparently rare in practice. 

\section{Computational  Comparison of Greedy, Optimal, and Rollout Policies}
In this section, we will present our computational studies of the proposed rollout approaches in two contexts. We will first consider small-scale Markov chains and $N=100$ steps, where computing the optimal policy via the DP-like algorithm discussed in Section~2 is feasible. Our goal is to demonstrate that the rollout algorithm and its variants produce $N$-step sequences whose probability of occurrence is close to the optimal. In contrast, the ones selected by the greedy policy are much less likely. 

We will then consider the Markov chain defined by a fine-tuned GPT, modified from the open source implementation given by Karpathy in [Kar22]. Due to the large size of state space, computing a most likely sequence from a given initial state is intractable. We will show that our rollout approaches are effective for this problem despite its scale, and that substantial improvements over the greedy policy are obtained.

\subsection{Small-Scale Markov Chains}
We will present the results of our experiments with small-scale Markov chains. The size of these chains is small enough so that the DP algorithm (or a Viterbi algorithm) can be used to compute the most likely sequence starting from every initial state. Thus, the performance differences between the rollout, greedy, and optimal policies can be accurately assessed. We describe how the Markov chains are generated so that they resemble those defined by a GPT, and we provide metrics according to which the performance of rollout is evaluated. 

We consider problems involving $100$ states, and we demonstrate that the performance improvement of rollout with one-step and multistep lookahead over the greedy policy is substantial, consistent with earlier experience with rollout algorithms and the Newton step conceptualization that underlies them. Similarly, we find that truncated rollout algorithms perform nearly as well as their untruncated counterparts, while requiring much less computation. Our experiments illustrate some typical patterns in occurrence probabilities of sequences computed via optimal, rollout, and greedy policies. We also present the performance of untruncated and truncated double rollout with one-step and multistep lookahead, and we note that the truncated versions remain effective. In addition, we find that the performance improvement of double rollout over (single) rollout is substantial, even when the number of lookahead steps is small.

Let us now describe the process through which the Markov chains have been generated. We assume that there is a fixed number $q$ of states $y$ such that $p(y\ |\ x)>0$, with $q$ being the same for all states $x$. In the context of an $n$-gram, where the state space $X$ is the set of all $n$-word sequences, $q$ is the vocabulary size, while the state space size, denoted by $|X|$, is the cardinality of $X$; clearly both of these numbers can be enormous. We refer to the ratio $q/|X|$ (in percent) as the \emph{branching factor} of the Markov chain. For each state $x\in X$, we generate, according to a uniform distribution, a set of $q$ distinct states $y$ such that $p(y\,|\,x)>0$. The probabilities $p(y\,|\,x)$ are also generated according to a uniform distribution. 

Given a Markov chain with state space $X$ and fixed branching factor as described above, we consider the most likely sequence selection problem with sequence length $N$, starting from every initial state. We compute the most likely sequence via the DP-like algorithm described in Section~2 and the sequence given by the greedy policy. They are used to evaluate the performance of our rollout approaches. The probability of an entire sequence is typically very small, so we will represent it as the average of its constituent transition probabilities (i.e., a geometric mean over $N$ as will be described below). 

In particular, given a sample set  $C$ of Markov chains, we compute the optimal occurrence probability of generated sequences,  averaged over all chains, states, and transitions, and denoted by $(P^*_0)^{1/N}$, according to the average geometric mean formula
$$(P^*_0)^{1/N}=\frac{\sum_{c\in C}\sum_{x\in X}\big(P_{0,c}^*(x)\big)^{1/N}}{|C|\cdot|X|},$$
where $P_{0,c}^*(x)$ is the optimal occurrence probability with $x_0=x$ and Markov chain $c$ in the sample set. Similarly, we compute the occurrence probabilities of sequences generated by the greedy policy averaged over all chains, states, and transitions, and denoted by $(\overline{P}_0)^{1/N}$, according to
\begin{equation}
    \label{eq:average_prob_greedy}
    (\overline{P}_0)^{1/N}=\frac{\sum_{c\in C}\sum_{x\in X}\big(P_{0,c}(x,\overline{\pi})\big)^{1/N}}{|C|\cdot|X|}.
\end{equation}
Here $P_{0,c}(x,\overline{\pi})$ is the transition probability of the sequence generated by the greedy policy with $x_0=x$ and Markov chain indexed by $c$. For the rollout algorithm (or variants thereof), we compute its averaged occurrence probability $(\Tilde{P}_0)^{1/N}$ similar to Eq.~\eqref{eq:average_prob_greedy} with $\Tilde{\pi}$ in place of $\overline{\pi}$. Then the performance of this rollout approach is measured by its {\it percentage  recovery} of optimality loss of the greedy policy, given by 
\begin{equation}
    \label{eq:metric_small_chain}
    \frac{(\Tilde{P}_0)^{1/N}-(\overline{P}_0)^{1/N}}{(P^*_0)^{1/N}-(\overline{P}_0)^{1/N}}\times 100\,\text{(\%)}.
\end{equation}
This performance measure describes accurately how the rollout performance compares with the greedy policy and how close it comes to optimality.

\subsection*{Experimental Results}

\begin{figure}[t]
\begin{center}
\centerline{{\includegraphics[width=\columnwidth]{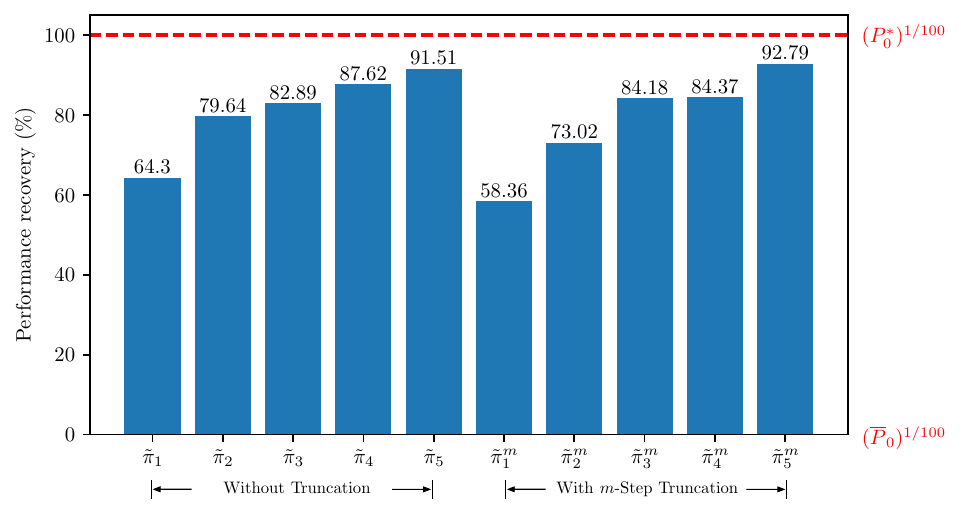}}}
\caption{Percentage recovery of the optimality loss of the greedy policy through the use of rollout and its variants, applied to sequence selection problems with $N=100$ for $50$ randomly generated Markov chains with $100$ states and $5\%$ branching factor. Here $\Tilde{\pi}_{\ell}$ represents rollout with $\ell$-step lookahead, and $\Tilde{\pi}_{\ell}^m$ represents $m$-step truncated rollout for $m=10$ with $\ell$-step lookahead. It can be seen that on average, rollout and its variants provide a substantial improvement over the greedy policy, that the improvement increases with the size of the lookahead, and that truncated rollout methods perform comparably to their exact counterparts.}
		\label{fig:rollout_small}
		\end{center}
\end{figure}

In our experiments the percentage recovery has ranged roughly from $60\%$ to $90\%$ for one-step to five-step lookahead, untruncated and truncated rollout with $m=10$ steps up to truncation; see  Fig.~\ref{fig:rollout_small}. The performance improves as the length of the lookahead increases, but seems remarkably unaffected by the $90\%$ truncation of the rollout horizon (the relative insensitivity of the performance of truncated rollout to the number of rollout steps $m$ has been observed in other application contexts as well).
The figure has been generated with a sample of $50$ different Markov chains with $|X|=100$ states, branching factor equal $5\%$, and sequence length $N=100$. We tested rollout with one-step and multistep lookahead (ranging from $2$ to $5$ steps), and their $m$-step truncated counterparts with $m=10$. Their percentage  recovery, evaluated according to Eq.~\eqref{eq:metric_small_chain}, is given in Fig.~\ref{fig:rollout_small}, where $\Tilde{\pi}_{\ell}$ denotes rollout with $\ell$-step lookahead, and $\Tilde{\pi}_{\ell}^m$ denotes $m$-step truncated rollout with $\ell$-step lookahead. 

It can be seen that the sequences produced by rollout improve substantially over those generated by the greedy policy. In fact, the sequences generated by the untruncated rollout policy $\Tilde{\pi}$ all have larger occurrence probabilities than those generated by the greedy policy, consistent with the analysis given in Section~4. In addition, the performance improves, on average, as the size of the lookahead increases. However, we have observed that this is not true for rare individual examples. We have also observed that there is only a small degradation of performance when applying the truncated rollout compared with untruncated rollout for all lookahead sizes considered. This is significant as truncated rollout greatly reduces the computation if $m\ll N$ (see its complexity analysis in Section~3). 

Fig.~\ref{fig:pattern_small} illustrates the typical patterns in average probabilities computed via the optimal, the rollout and its $2$- and $3$-step lookahead variants, and the greedy policies. Given a Markov chain, we compute the probabilities $\big(P_0(x,\pi)\big)^{1/N}$ for each state $x\in X$, where $\pi$ can be $\pi^*$, $\Tilde{\pi}_\ell$ with $\ell=1,2,3$, or $\overline{\pi}$. Each sub-figure in Fig.~\ref{fig:pattern_small} illustrates the results corresponding to a single representative Markov chain. However, different probability patterns do not appear in equal proportions in our numerical experiments. In particular, the pattern that appears in Fig.~\ref{fig:pattern_small_2} is relatively rare. 

Generally, the common feature shared by all the results in  Fig.~\ref{fig:pattern_small} is that rollout and its variants result in substantial improvement over the greedy policy across all states. Moreover, longer lookahead leads to more likely sequences for the great majority of initial states. 
However, there are some notable differences in the patterns shown in  Fig.~\ref{fig:pattern_small}. In Fig.~\ref{fig:pattern_small_1}, longer lookahead (up to 3) does not produce significant improvement over one-step lookahead. Similarly, in Fig.~\ref{fig:pattern_small_2}, sequences selected by rollout with one-step lookahead are already near optimal. A relatively rare phenomenon shown in Fig.~\ref{fig:pattern_small_2} is that the longer lookahead with $\ell=3$ deteriorates the performance of rollout for many states. Figs.~\ref{fig:pattern_small_3} and \ref{fig:pattern_small_4} represent fairly common patterns, where longer lookahead results in substantial improvement.
\begin{figure}[p]
    \centering
    \begin{subfigure}{\columnwidth}
        \centering
        \includegraphics[width=\linewidth]{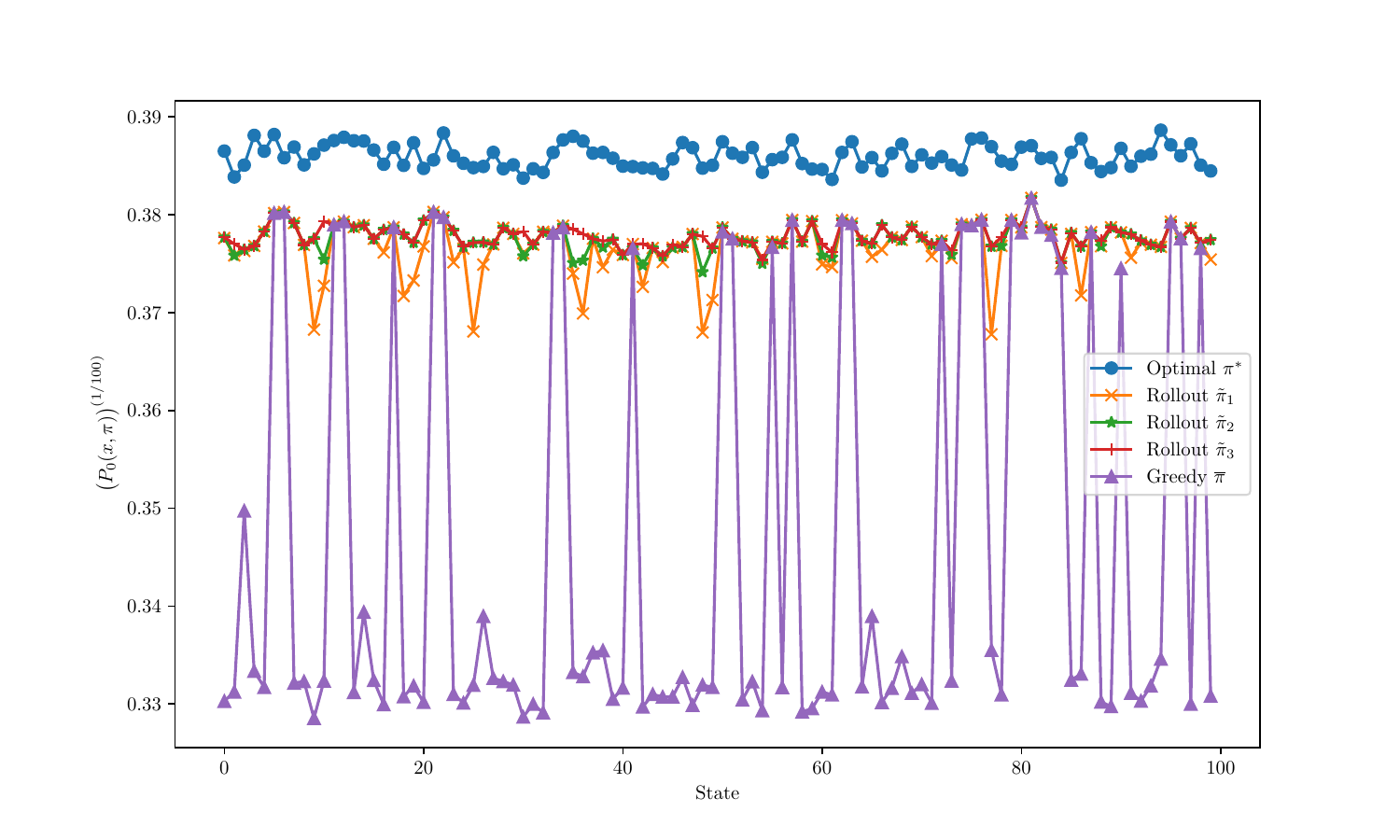}
        \caption{Problem where longer lookahead (up to 3) does not result in significant improvement over one-step lookahead.}
        \label{fig:pattern_small_1}
    \end{subfigure}
    \begin{subfigure}{\columnwidth}
        \centering
        \includegraphics[width=\linewidth]{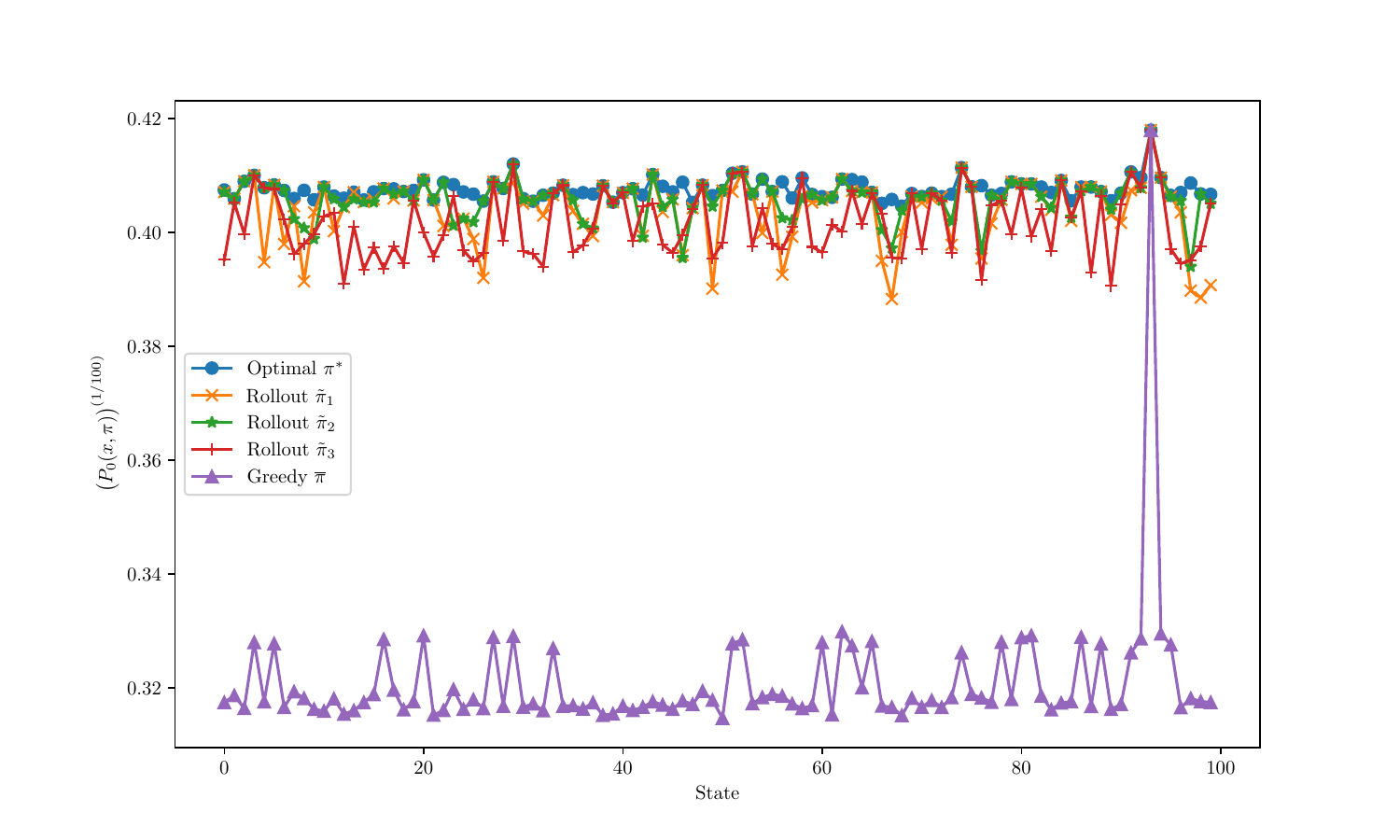}
        \caption{Problem where one-step lookahead produces highly likely sequences.}
        \label{fig:pattern_small_2}
    \end{subfigure}
    \caption{Typical patterns in average probabilities computed via the optimal, rollout and its $2$- and $3$-step lookahead variants, and the greedy policies. Given a Markov chain and for each state $x\in X$, we compute the probabilities $\big(P_0(x,\pi)\big)^{1/N}$, where $\pi$ can be $\pi^*$, $\Tilde{\pi}_\ell$ with $\ell=1,2,3$, and $\overline{\pi}$. Substantial improvements are obtained from rollout for all states.}
\end{figure}
\begin{figure}[p]\ContinuedFloat
    \centering
    \begin{subfigure}{\columnwidth}
        \centering
        \includegraphics[width=\linewidth]{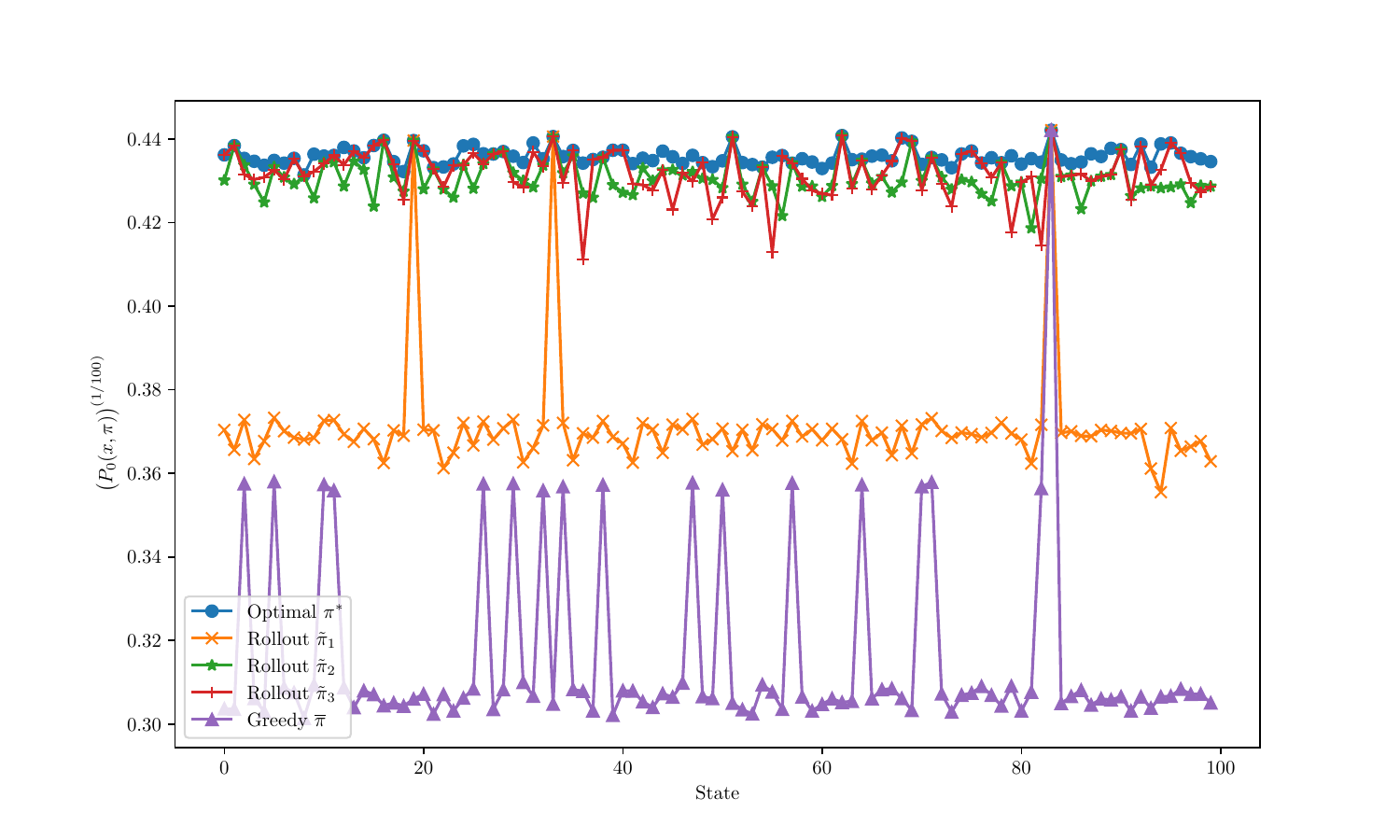}
        \caption{Problem where large improvement occurs in both the first and second lookahead steps.}
        \label{fig:pattern_small_3}
    \end{subfigure}
    \begin{subfigure}{\columnwidth}
        \centering
        \includegraphics[width=\linewidth]{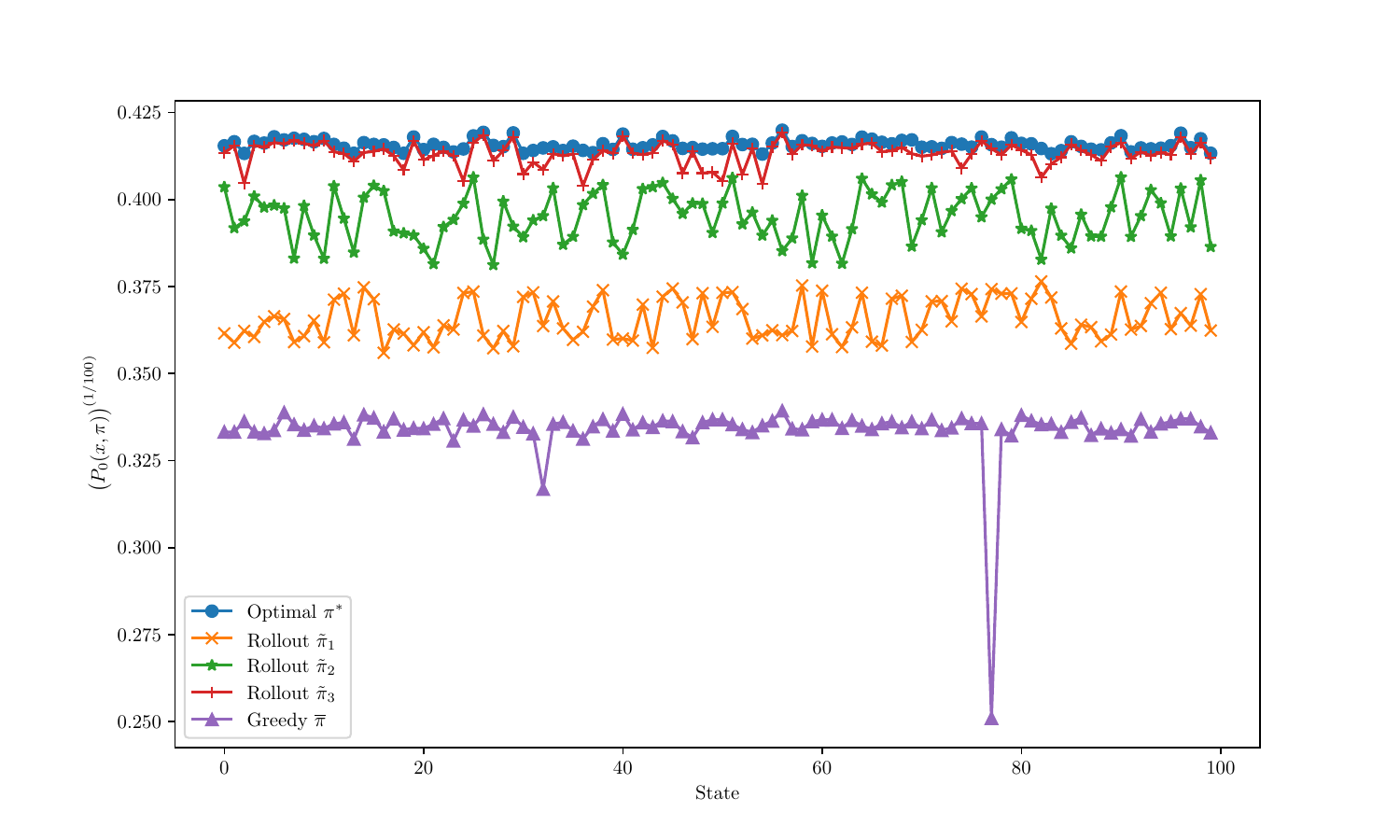}
        \caption{Problem where longer lookahead leads to steady and gradual improvement.}
        \label{fig:pattern_small_4}
    \end{subfigure}
    \caption{Typical patterns in average probabilities computed via the optimal, rollout and its $2$- and $3$-step lookahead variants, and the greedy policies. Given a Markov chain and for each state $x\in X$, we compute the probabilities $\big(P_0(x,\pi)\big)^{1/N}$, where $\pi$ can be $\pi^*$, $\Tilde{\pi}_\ell$ with $\ell=1,2,3$, and $\overline{\pi}$. Substantial improvements are obtained from rollout for all states.}
    \label{fig:pattern_small}
\end{figure}

\subsection*{Double Rollout}

We will now provide additional results using double rollout. This is the rollout method discussed in Section 3, with its base policy given by a rollout method as well. In particular, we use as the base policy the one-step lookahead untruncated rollout policy $\Tilde{\pi}_1$, cf. Fig.~\ref{fig:rollout_small}. This double rollout method can be viewed as two successive policy iterations, as discussed in Section~3. Note that double rollout is applied in real-time and computes only for states that are needed, in contrast with conventional applications of policy iteration, which operate off-line and for all states. Note also that other rollout variants can be used as base policies, such as truncated and/or multistep lookahead rollout.

\begin{figure}[t]
\begin{center}
\centerline{{\includegraphics[width=\columnwidth]{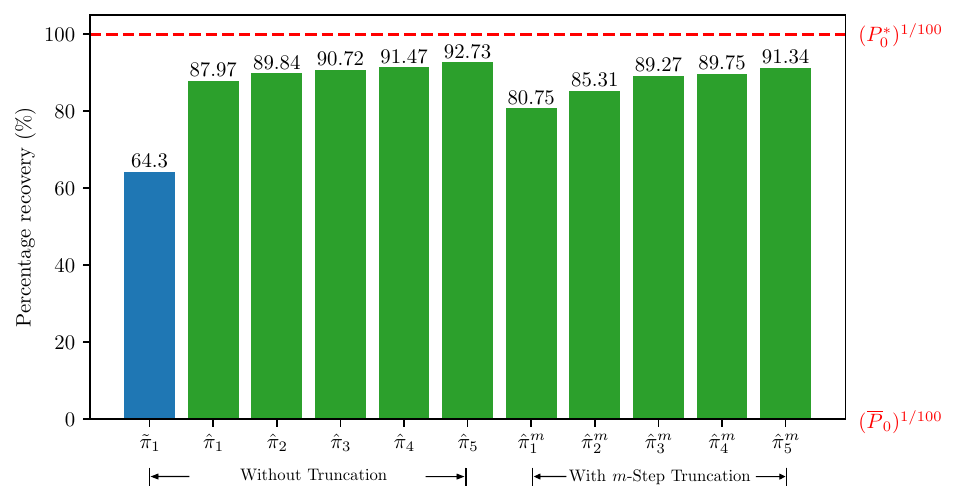}}}
\caption{Percentage recovery of the optimality loss of the greedy policy through the use of double rollout and its variants, applied to sequence selection problems with $N=100$ for $50$ randomly generated Markov chains with $100$ states and $5\%$ branching factor. Here $\Tilde{\pi}_1$ (in blue) represents (single) rollout with one-step lookahead, $\hat{\pi}_{\ell}$ represents double rollout with $\ell$-step lookahead, and $\hat{\pi}_{\ell}^m$ represents $m$-step truncated double rollout for $m=10$ with $\ell$-step lookahead.}
		\label{fig:double_rollout_small}
		\end{center}
\end{figure}

We have tested untruncated and truncated double rollout with one-step and multistep lookahead (ranging from $2$ to $5$ steps) using the same set of problems as for Fig.~\ref{fig:rollout_small}. The percentage recovery of double rollout and its variants over the greedy policy are shown as green bars in Fig.~\ref{fig:double_rollout_small}, where $\hat{\pi}_{\ell}$ is double rollout with $\ell$-step lookahead, and $\hat{\pi}_{\ell}^m$ represents $m$-step truncated double rollout for $m=10$ with $\ell$-step lookahead. We also include the result for rollout with one-step lookahead, denoted by $\Tilde{\pi}_1$ and shown as a blue bar, for easy comparison. It can be seen that the double rollout algorithm and its variants lead to significant performance improvement, not only over the greedy policy, but also over (single) rollout with one-step lookahead. Note that the truncated versions of double rollout remain effective, despite large computational savings.

Let us also comment briefly on our test results for problems involving $1000$ states, $1\%$ branching factor, and sequence length $N=1000$. The performance improvement of the rollout algorithms over the greedy policy is qualitatively similar to those shown in Figs.~\ref{fig:rollout_small}, \ref{fig:pattern_small}, and \ref{fig:double_rollout_small}. In the case of truncated rollout we have used $m=10$, so that at each state the method simulates only $1\%$ of the remaining sequence.

\subsection{The Markov Chain Defined by a GPT}
We will now discuss computational experimentation with our methods applied to text generation with a GPT programmed by Karpathy [Kar22], which we fine-tuned in a way to be  described shortly. In this context, it is intractable to compute the most likely sequence via the DP-like algorithm described in Section~2. Moreover, the large vocabulary size can also lead to excessive computation when rollout is applied.  For these reasons, we used simplified rollout  with one-step lookahead, and its truncated counterpart. We have also taken advantage  of graphical processing units (GPU); the Q-factors of 
our rollout schemes can be computed in parallel at each state.

In particular, a GPT with fixed weights defines an $n$-gram and corresponding Markov chain along with its transition probabilities.\footnote{After training or fine-tuning a GPT, the output probabilities of the GPT may be further modified through tuning of some additional parameters that are internal to the GPT. For example, to suppress the repeated content generated by the model, penalty terms that are absent in the training phase are introduced in ChatGPT [Ope24]. These parameters may be introduced for various reasons. However, regardless of their practical purpose, once these parameters are selected, the Markov chain and its transition probabilities are well-defined.} For this Markov chain, the size of the state space is $q^n$, where $q$ is the vocabulary size. Since both $q$ and $n$ are large in modern language models, computing the most likely sequence from a given initial word string is intractable, even for a small sequence length. The GPT used in our computational studies is built upon that provided in [Kar22], which involves 124 million weights. The initial values of the weights are those given in [RWC19]. With this GPT, the values of $n$ and $q$ are $1024$ and $50258$, respectively, and we aim to compute word sequences with length $N=200$. We fine-tuned the GPT with a dataset composed of the writings of James A. Michener. 

\begin{figure}[t]
\begin{center}
\centerline{{\includegraphics[width=\columnwidth]{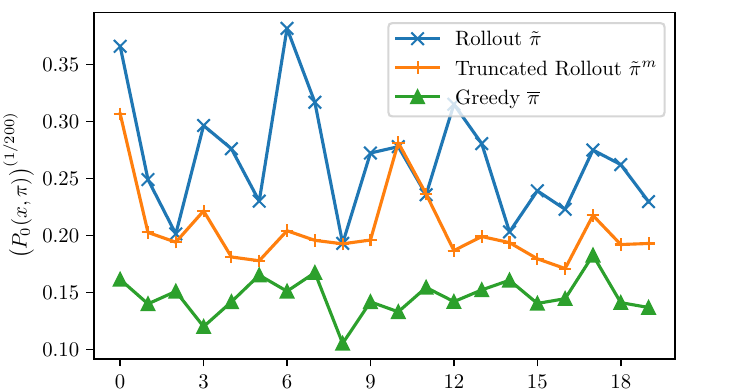}}}
\caption{Probabilities of occurrence of sequences generated by the greedy policy, and the truncated and untruncated simplified rollout policies with one-step lookahead. At each step, the simplified rollout policy computes $10$ Q-factors, corresponding to the top ten most likely next words. The $m$-step truncated rollout simulates 5\% of the remaining sequence. The figure shows the results for 20 different initial states.}
		\label{fig:simplified_rollout}
		\end{center}
\end{figure}

We generated 20 different initial states $x_0$ using GPT4 with the prompt ``Provide twenty opening sentences for James A. Michener style fiction." The GPT4 responses were padded with place-holder words to the proper length, to form initial states. For each initial state $x_0$, we applied both the greedy and the simplified rollout policies for sequence selection. At each step, the simplified rollout policy computes $10$ Q-factors, corresponding to the top ten most likely next words. The $m$-step truncated rollout was implemented with $m=10$, so that it simulates $m/N=5\%$ of the remaining sequence. Fig.~\ref{fig:simplified_rollout} shows the probabilities of occurrence of sequences generated by the greedy policy, and the truncated and untruncated simplified rollout policies, for the 20 different initial states. It can be seen from this figure that substantial performance improvements have been obtained by the untruncated and $m$-step truncated simplified rollout policies over that of the greedy policy in all of our 20 test cases, consistent with the analysis of Section~4, as well as the computational results given earlier for the small-scale Markov chains.

\section{Concluding Remarks}

We have proposed new algorithms for finding highly likely sequences in Markov chains and their applications in $n$-grams, transformers, and HMMs. The algorithms are based on the rollout approach, which is a single policy iteration starting from a base policy. In our case the base policy was chosen to be the greedy policy, which is natural in the context of Markov chains. However, other base policies may be convenient for use in specialized contexts. Similar to the present paper, application of rollout will often lead to substantial performance improvement over these base policies at modest computational cost. 

There are also constrained variants of the problem of this paper, whereby the generated $N$-state sequence must satisfy some constraints (for example some states or cycles of states cannot be repeated). Rollout algorithms can be adapted to this context; see the textbook literature cited earlier. Another interesting case relates to Markov chains with a termination state. Then the problem is to maximize the probability of occurrence among sequences that start at a given initial state and end at the termination state. The methodology of the present paper can also be adapted for this type of problem, but we have not done any related experiments.

Finally let us note that in the context of transformers of the ChatGPT type, it is not clear whether and in what situations obtaining a highly likely generated sequence will lead to qualitative improvement of the results obtained. A relevant issue here is that there may not exist a clear and objective measure of quality of the transformer output. This is a subject of much current interest.

\def\refspace{\par\noindent}
\section*{References}

\refspace[BTW97] Bertsekas, D.\ P., Tsitsiklis, J.\ N., and Wu, C., 1997.\ ``Rollout Algorithms for
Combinatorial Optimization,'' Heuristics, Vol.\ 3, pp.\ 245-262.

\refspace[Ber17] Bertsekas, D.\ P., 2017.\ Dynamic Programming and Optimal Control,  Vol.\ I, Athena Scientific, Belmont, MA.

\refspace[Ber19] Bertsekas, D.\ P., 2019.\ Reinforcement Learning and Optimal Control,  Athena Scientific, Belmont, MA.

\refspace[Ber20] Bertsekas, D.\ P., 2020.\ Rollout, Policy Iteration, and Distributed Reinforcement Learning,  Athena Scientific, Belmont, MA.

\refspace[Ber22] Bertsekas, D.\ P., 2022.\ Lessons from AlphaZero for Optimal, Model Predictive, and Adaptive Control,  Athena Scientific, Belmont, MA.

\refspace[Ber23] Bertsekas, D.\ P., 2023.\ A Course in Reinforcement Learning,  Athena Scientific, Belmont, MA.

\refspace[BiB24] Bishop, C.\ M, and Bishop, H., 2024.\ Deep Learning: Foundations and Concepts,
Springer, New York, N.\ Y.

\refspace[DEK98]  Durbin, R., Eddy, S.\ R., Krogh, A., and Mitchison, G., 1998.\ Biological Sequence Analysis, Cambridge Univ.\ Press, Cambridge.

\refspace[Edd96] Eddy, S.\ R., 1996.\ ``Hidden Markov Models," Current Opinion in Structural Biology, Vol.\ 6, pp.\ 361-365.

\refspace[For73] Forney, G.\ D., 1973.\ ``The Viterbi Algorithm," Proc. IEEE, Vol.\
61, pp.\ 268-278.

\refspace[JuM23] Jurafsky, D., and Martin, J.\ H., 2023.\ Speech and Language Processing: An Introduction to Natural Language Processing, Computational Linguistics, and Speech Recognition, draft 3rd edition (on-line).

\refspace[Kar22] Karpathy, A., 2022.\ nanoGPT, \url{https://github.com/karpathy/nanoGPT}.

\refspace[MaS99] Manning, C., and Schutze, H., 1999.\ Foundations of Statistical Natural Language Processing, MIT Press, Cambridge, MA.

\refspace[Ope24] OpenAI, 2024.\ Text Generation Models, \url{https://platform.openai.com/docs/guides/text-generation}.

\refspace[PrS01] Proakis, J.\ G., and Salehi, M., 2001.\  Communication Systems
Engineering, Prentice-Hall, Englewood Cliffs, N.\ J.

\refspace[PrS08] Proakis, J.\ G., and Salehi, M., 2008.\  Digital Communications, McGraw-Hill, N.\ Y.

\refspace[RWC19] Radford, A., Wu, J., Child, R., Luan, D., Amodei, D. and Sutskever, I., 2019.\ ``Language Models Are Unsupervised Multitask Learners," OpenAI Blog, 1(8), pp.\ 1-24.

\refspace[Rab89]  Rabiner, L.\ R., 1989.\  ``A Tutorial on Hidden Markov Models and Selected Applications in
Speech Recognition,'' Proc.\ of the IEEE, Vol.\ 77, pp.\
257-286.

\refspace[Vit67] Viterbi, A.\ J., 1967. ``Error Bounds for Convolutional Codes and an
Asymptotically Optimum Decoding Algorithm," IEEE Trans.\ on Info.\ Theory, Vol.\
IT-13, pp.\ 260-269.

\refspace[YDR04] Yan, X., Diaconis, P., Rusmevichientong, P., and Van Roy, B., 2004.\
``Solitaire: Man Versus Machine,'' Advances in Neural Information
Processing Systems, Vol.\ 17, pp.\ 1553-1560.

\end{document}